%% file: iclr2023_conference.tex
\documentclass{article} 
\usepackage{iclr2023_conference,times}

\input{math_commands.tex}

\usepackage{hyperref}
\usepackage{url}
\usepackage{booktabs}
\usepackage{graphicx}
\usepackage{csquotes}
\usepackage{color, soul}
\usepackage{fancyhdr}

\graphicspath{ {./images/} }

\newcommand{\squeezeup}{\vspace{-2.5mm}}

\title{N2G: A scalable approach for quantifying interpretable neuron representations in Large Language Models}

\author{Alex Foote$^{1}$\thanks{Equal Contribution}, Neel Nanda$^{2}$, Esben Kran$^{1}$, Ionnis Konstas$^{3}$, Fazl Barez$^{1, 3, 4}$\footnotemark[\value{footnote}]\\  
$^{1}$Apart Research \ $^{2}$Independent \ $^{3}$Edinburgh Centre for Robotics \
$^{4}$ University of Oxford 
}

\hypersetup{
    colorlinks,
    citecolor=[rgb]{0.151, 0.2, 0.651},
    linkcolor=[rgb]{0.151, 0.2, 0.651},
    urlcolor=[rgb]{0.151, 0.2, 0.651},
}

\iclrfinalcopy 
\begin{document}

\maketitle

\begin{abstract}
Understanding the function of individual neurons within language models is essential for mechanistic interpretability research. 
We propose \textbf{Neuron to Graph (N2G)}, a tool which takes a neuron and its dataset examples, and automatically distills the neuron's behaviour on those examples to an interpretable graph. This presents a less labour intensive approach to interpreting neurons than current manual methods, that will better scale these methods to Large Language Models (LLMs). We use truncation and saliency methods to only present the important tokens, and augment the dataset examples with more diverse samples to better capture the extent of neuron behaviour. These graphs can be visualised to aid manual interpretation by researchers, but can also output token activations on text to compare to the neuron's ground truth activations for automatic validation. 
N2G represents a step towards scalable interpretability methods by allowing us to convert neurons in an LLM to interpretable representations of measurable quality.
\end{abstract}

\section{Introduction}

Interpretability of machine learning models is an active research topic \citep{hendrycks2021unsolved, amodei2016concrete} and can have a wide range of applications from bias detection \citep{vig_investigating_2020}  to autonomous vehicles \citep{barez2022system} and Large Language Models (LLMs; \citet{elhage2022solu}). The growing subfield of mechanistic interpretability aims to understand the behaviour of individual neurons within models as well as how they combine into larger circuits of neurons that perform a particular function \citep{olah2020zoom, olah2022mechanistic, goh2021multimodal}, with the ultimate aim of decomposing a model into interpretable components and using this to ensure model safety. 

Interpretability tools for understanding neuron in LLMs are lacking. Currently, researchers often look at dataset examples containing tokens on which a neuron strongly activates and investigate common elements and themes across examples to give some insight into neuron behaviour \citep{elhage2022solu, geva2020transformer}. However, this can give the illusion of interpretability when real behaviour is more complex \citep{bolukbasi2021interpretability}, and measuring the degree to which these insights are correct is challenging. Additionally, inspecting individual neurons by hand is time-consuming and unlikely to scale to entire models. 

To overcome these challenges, we present \textbf{Neuron to Graph (N2G)}, which automatically converts a target neuron within an LLM to an interpretable graph that visualises the contexts in which the neuron activates.
The graph https://www.overleaf.com/project/641173ac7cb539e827754acbcan be visualised to facilitate understanding the neuron's behaviour, as well as used to process text and produce predicted token activations. This allows us to measure the correspondence between the target neuron's activations and the graph's activations, which provides a direct measurement of the degree to which a graph captures the neuron's behaviour. 

Our method takes maximally activating dataset examples for a target neuron, prunes them to remove irrelevant context, identifies the tokens which are important for neuron activation, and creates additional examples by replacing the important tokens with other likely substitutes using BERT \citep{bert}. These processed examples are then given as input to the graph builder, which removes unimportant tokens and creates a condensed representation in the form of a trie. This trie can then be used to process text and will predict activations for each token, and can be converted to a graph for visualisation.


\squeezeup
\begin{figure}[ht]
\centering
\includegraphics[width=13.5cm]{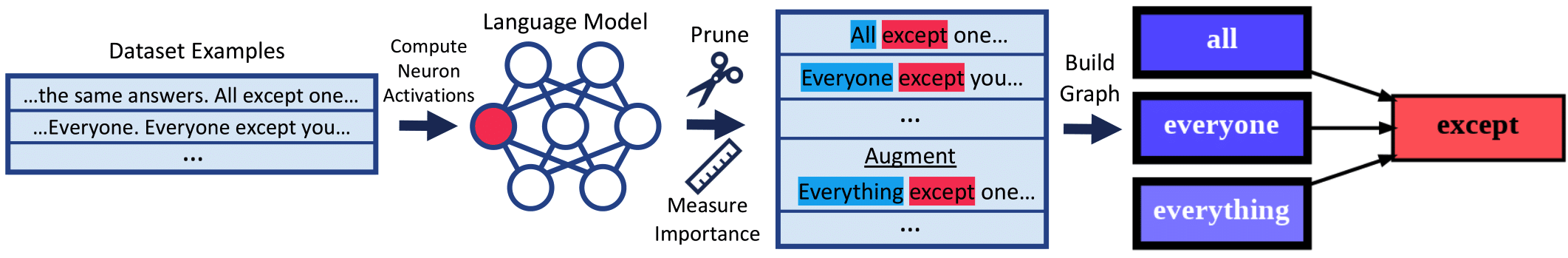}
\label{figure:architecture}
\caption{\textbf{Overall architecture of N2G.} Activations of the target neuron on the dataset examples are retrieved (neuron and activating tokens in red). Prompts are pruned and the importance of each token for neuron activation is measured (important tokens in blue). Pruned prompts are augmented by replacing important tokens with high-probability substitutes using BERT. The augmented set of prompts are converted to a graph. The output graph is a real example which activates on the token \textquote{except} when preceded by any of the other tokens.}
\end{figure}
\squeezeup
\squeezeup

\begin{figure}[ht]
\centering
\includegraphics[width=11cm]{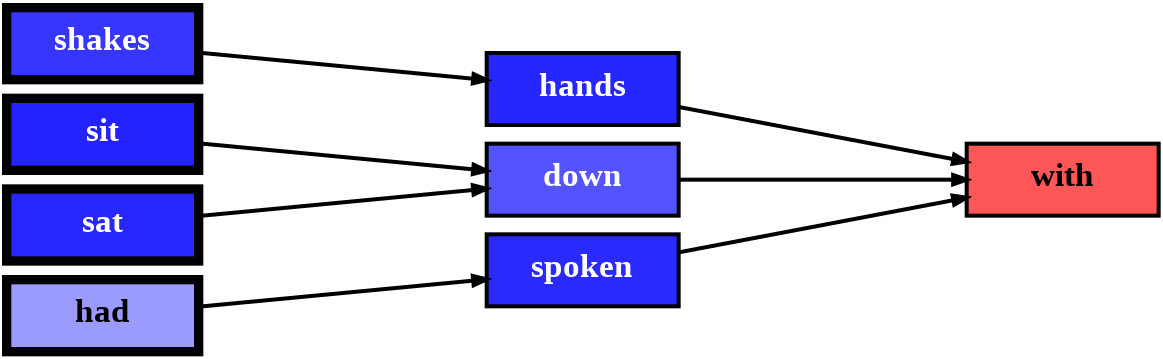}
\caption{An example of a graph built from Neuron 2 of Layer 1 of the model.}
\label{figure:graph_example}
\end{figure}
\squeezeup
\squeezeup


\section{Related Work}


Prior work in neuron analysis has identified the presence of neurons correlated with specific concepts \citep{radford2017learning}. For instance, \citet{Dalvi2019} explored neurons which specialised in linguistic and non-linguistic concepts in large language models, and \citet{seyffarth-etal-2021-implicit} evaluated neurons which handle concepts such as causation in language. The existence of similar concepts embedded within models can also be found across different architectures. \citet{wu-etal-2020-similarity} and \citet{schubert2021high} examined neuron distributions across models and found that different architectures have similar localised representations of information, even when \citet{durrani-etal-2020-analyzing} used a combination of neuron analysis and visualization techniques to compare transformer and recurrent models, finding that the transformer produces fewer neurons but exhibits stronger dynamics.

There are various methods of identifying concept neurons \citep{geva2020transformer}. \citet{bau-et-al-2018} proposed a method of identifying important neurons across models by analyzing correlations between neurons from different models. In contrast, \citet{dai-et-al-2021} developed a method to identify concept neurons in transformer feed-forward networks by computing the contribution of each neuron to the knowledge prediction. 
In contrast, we focus on identifying neurons using highly activating dataset examples. \citet{mu-and-andreas-2020} demonstrated how the co-variance of neuron activations on a dataset can be used to distinguish neurons that are related to a particular concept. \citet{torroba-hennigen-etal-2020-intrinsic} also used neuron activations to train a probe which automatically evaluates language models for neurons correlated to linguistic concepts.

One limitation of using highly activating dataset examples is that the accurate identification of concepts correlated with a neuron is limited by the dataset itself. A neuron may represent several concepts, and  \citet{Bolukbasi-et-al-2021} emphasise the importance of conducting interpretability research on varied datasets, in order to avoid the “interpretability illusion”, in which neurons that show consistent patterns of activation in one dataset activate on different concepts in another. \citet{poerner-etal-2018-interpretable} also showed the limitations of datasets in concept neuron identification. They demonstrated that generating synthetic language inputs that maximise the activations of a neuron surpasses naive search on a corpus.
\section{Methodology}

N2G constructs an interpretable graph representing the context required for a given neuron to activate on a particular token. The graph can be used to process text and predict whether the target neuron will fire (strongly activate) for each token in the input. N2G takes as input a model, the layer and neuron indices of the target neuron, and a set of prompts which contain one or more tokens for which the target neuron strongly activates. Figure \ref{figure:architecture} illustrates the overall process as well as an example of a real graph. Figure \ref{figure:graph_example} shows another example and Appendix \ref{appendix:graphs} contains further examples with interesting behaviours. 


\textbf{Prune}: Given a prompt, we process it with the model and retrieve the token activations of the target neuron using the TransformerLens library \citep{nandatransformerlens2022}, which allows easy access to internal neuron activations. The prune function takes the prompt and activations and finds the token with the highest activation (the key token). It removes all sentences after the key token (we study autoregressive models so these cannot affect neuron activation) and removes all tokens before the key token. It then measures the activation of the neuron on the key token in the truncated prompt to determine the change in activation. If this activation has decreased by more than a user-defined percentage (we choose $50\%$), then the prior token is added to the truncated prompt. This process is then repeated until the neuron activation on the key token is sufficient to pass the condition.

\textbf{Saliency}: The importance of each token for neuron activation on every other token is then computed to create a matrix of token importance. The importance $I_k$ of the $k^{th}$ token relative to the $j^{th}$ token is calculated as $I_k=1 - (a_{j,\textit{masked}}/a_j)$, where $a_j$ is the activation of the neuron on the $j^{th}$ token and $a_{j,\textit{masked}}$ is the activation of the neuron on the $j^{th}$ token when token $k$ is masked with a special padding token. This method is similar to other perturbation-based saliency methods in Computer Vision \citep{Dabkowski2017RealTI} and NLP \citep{liu2018nlize}. 

\textbf{Augment}: The pruned prompt is then used to generate more varied inputs to better explore the neuron's behaviour. Each token that is important for activation on the key token is masked in turn, and BERT \citep{bert} predicts the top $n$ substitutions for the masked token. A new prompt is created for each substitute token, provided they cross a probability threshold. This technique is very similar to existing methods of data augmentation used during training \citep{ma2019nlpaug}.

\textbf{Graph Building}: The pruned prompts and the augmented prompts are the input to the graph-building stage, along with normalised token activations and the importance matrix for each prompt. The normalised activation $a_N$ of the $i^{th}$ token is calculated as $a_N=a_i/a_{\textit{max}}$, where $a_{\textit{max}}$ is the maximum activation of the neuron on any token in the training dataset. 

Each neuron graph is implemented as a trie, with each node representing a token. The first layer of nodes contains tokens on which the neuron strongly activates, and each sub-trie of one of these activating nodes represents the contexts for which the neuron will activate on that token. 

For every token in the prompt with a normalised activation above a threshold, we create a top layer node in the trie. Starting at the given activating token, we work backwards through the preceding tokens, adding them to the trie if they have an importance for the activating token above a threshold or adding them as a special ignore node if the importance is below the threshold. When processing text, ignore nodes are allowed to match to any token. Experimentally, we found that a normalised activation threshold of $0.5$ and an importance threshold of $0.75$ worked well. The final important token is marked as a termination node, which represents a valid stopping point when processing text. It records the normalised activation of the activating node for this path in the trie. We repeat this process for all activating tokens in all the input prompts. 


\textbf{Text Processing}: The resulting trie can be used to process text by beginning at the root of the trie and working backwards through the text prompt, checking if any consecutive sequence of prior tokens matches any path through the trie and reaches a termination node. We collate all valid matching paths and return the stored normalised activation on the longest matching path.

\textbf{Visualisation}: To visualize the trie, we create a condensed graph representation. We remove ignore nodes and termination nodes and create a layered graph by de-duplicating nodes by token value at each depth in the trie. We then color the activating nodes in the graph according to their normalised activation, with stronger activations corresponding to brighter red. Similarly, we color the rest of the token nodes in blue according to their importance. Additionally, we indicate nodes connected to a termination node in the full trie with a bold outline.

\section{Results and Discussion}
\label{section:results}

As the neuron graphs that are built by the algorithm can be directly used to process text and predict token activations, we can evaluate the degree to which they accurately capture the target neuron's behaviour by measuring the correspondence between the activations of the neuron and the predicted activations of the graph on some evaluation text. In our experiments we use a six-layer decoder-only Transformer model with SoLU activation, which may improve model interpretability by reducing polysemanticity \cite{elhage2022solu}. The model is trained on the Pile \citep{pile}, and we use data from Neuroscope \citep{nandaneuroscope2022}, which provides token level activations for the top $20$ prompts in the model's training set with the highest neuron activation on any token within the prompt, for all neurons in the model.

For each neuron, we take these top $20$ dataset examples and randomly split them in half to form a train and test set, and give the training examples to N2G to create a neuron graph. We then take the test examples and normalise the token activations as described above. We apply a threshold to the token activations, defining an activation above the threshold as a \textit{firing} of the neuron, and an activation below the threshold as the neuron \textit{not firing}. In these experiments we set the threshold to $0.5$.
We then process the test prompts with the neuron graph to produce predicted token firings. We can then measure the precision, recall, and $F1$ score of the graph's predictions compared to the ground truth firings.

Table \ref{table:avg_stats} shows the average precision, recall, and $F1$ score of the neuron graphs for a random sample of $50$ neurons from each layer of the model, stratified by neuron firing. In layer 0, the graphs on average capture the behaviour of the neurons well, with high recall and good precision on the tokens for which the real neuron fires, whilst maintaining near-perfect recall on the much larger number of tokens for which the neuron does not fire. Note that predicting token-level firings is in general a very imbalanced problem, as neurons typically fire on a small proportion of tokens in the input prompts. 

However, as we progress to deeper layers of the model, the recall and precision of the graphs generally decrease. This corresponds to neurons in the later layers on average exhibiting more complex behaviour that is less completely captured in the training examples. Specifically, neurons in early layers tend to respond to a small number of specific tokens in specific, narrow contexts, whereas later layers often respond to more abstract concepts represented by a wider array of tokens in many different contexts, which was similarly observed by \citep{elhage2022solu}. Precision also drops as the graphs may over-generalise and fail to capture the nuances of the context which caused a neuron to activate on a given token. 

\begin{table}
\centering


\begin{tabular}{ccccccc}
        \toprule
            & & \textbf{Firing Tokens} &                 &             & \textbf{Non-Firing Tokens} &                              \\
\textbf{Layer} & \textbf{Precision}  & \textbf{Recall} & \textbf{F1} & \textbf{Precision}      & \textbf{Recall} & \textbf{F1} \\
\midrule
0              & 0.74                & 0.85            & 0.74        & 1.0                     & 0.99            & 1.0         \\
1              & 0.66                & 0.77            & 0.64        & 1.0                     & 1.0             & 1.0         \\
2              & 0.60                & 0.77            & 0.6         & 1.0                     & 1.0             & 1.0         \\
3              & 0.48                & 0.70            & 0.48        & 1.0                     & 0.99            & 1.0         \\
4              & 0.44                & 0.72            & 0.46        & 1.0                     & 0.99            & 1.0         \\
5              & 0.45                & 0.67            & 0.42        & 1.0                     & 0.99            & 1.0         \\ \bottomrule
\end{tabular}
\label{table:avg_stats}
\caption{Precision, recall and $F1$-score of the neuron graphs' token-level predictions of neuron firing compared to ground truth on held-out test data, for 50 random neurons from each layer of the model. Tokens on which the real neuron fired and tokens on which it didn't fire are evaluated separately as there are generally many more tokens on which a neuron didn't fire, making it trivially easy to get near-perfect scores by always predicting the neuron will not fire.}

\end{table}

\begin{figure}[ht]
\centering
\includegraphics[width=13.8cm]{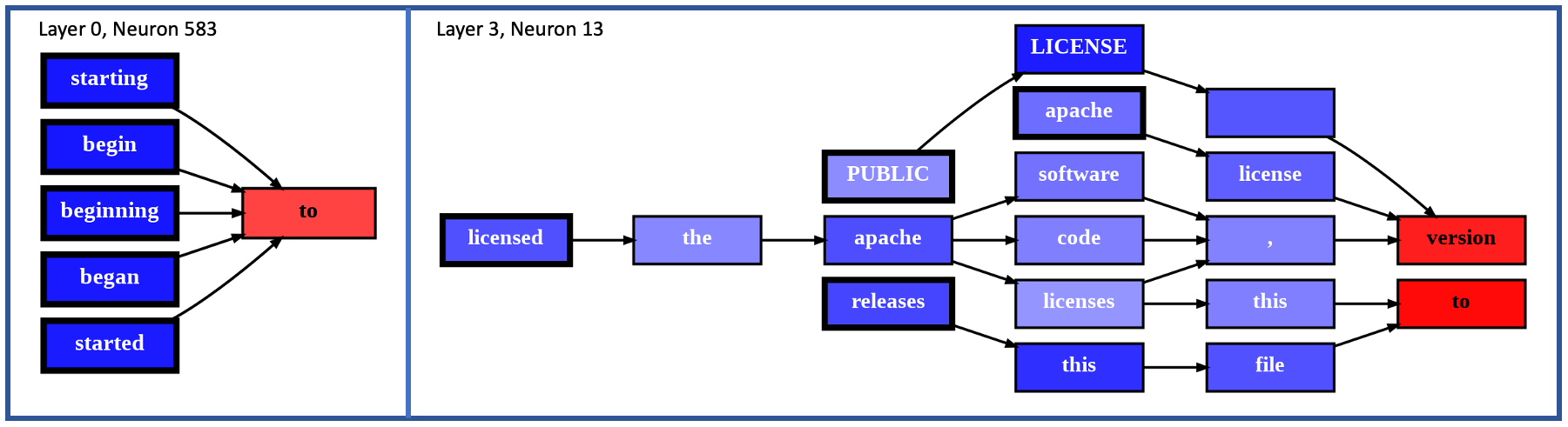}
\caption{Left: a graph from Layer 0 of the model. Right: a graph from Layer 3 of the model.}
\label{figure:comparison}
\end{figure}

For example, Figure \ref{figure:comparison} shows a comparison between a graph from Layer 0 and Layer 3. The graph from Layer 0 is typical for that layer - a small number of activating nodes that activate in simple contexts, often requiring just one of a small set of possible prior tokens to be present, and sometimes requiring no additional context at all. In contrast, the graph from Layer 3 exhibits a more complex structure, with longer and more intricate context that captures the more abstract concept of software licensing required for activation on the activating nodes.

\section{Conclusions and Limitations}




We introduced N2G, a method for automatically converting neurons in LLMs into interpretable graphs which can be visualized. The degree to which a graph captures the behaviour of a target neuron can be directly measured by comparing the output of the graph to the activations of the neuron, making this method a step towards scalable interpretability methods for LLMs. We find the neuron graphs capture neuron behaviour well for early layers of the model, but only partially capture the behaviour for later layers due to increasingly complex neurion behaviour, and this problem would likely become more prominent in larger models. Our approach primarily used SoLU \citep{elhage2022solu} models to reduce polysemanticity. Although also applicable to models with typical activation functions, the resulting graphs may need to be more comprehensive due to more complex neuron behaviours.
Our study focused on predicting neuron behaviour on the text that most activates it, excluding weaker activations. Future work could address these limitations by utilizing more training examples, better exploring the input space, and generalizing from exact token matches to matching abstract concepts, for example by using embeddings.



\bibliography{iclr2023_conference}
\bibliographystyle{iclr2021_conference}

\appendix
\section{Appendix}

\subsection{Neuron Graphs}
\label{appendix:graphs}

\begin{figure}[ht]
\centering
\includegraphics[width=13cm]{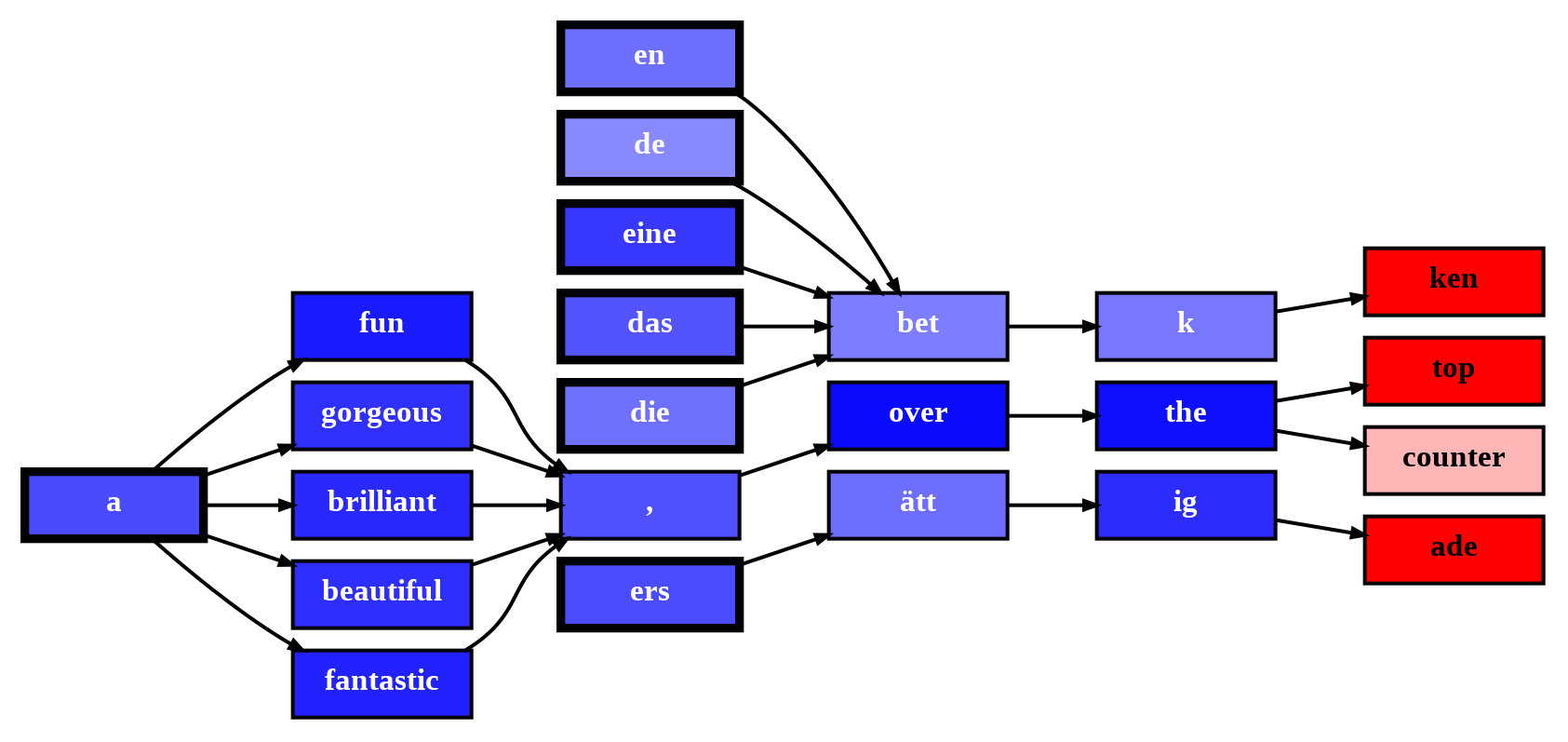}
\caption{A neuron graph exhibiting polysemanticity, with three disconnected subgraphs each responding to a phrase in a different language.}
\label{figure:polysemantic}
\end{figure}

In this section we explore some more interesting or characteristic behaviours of the neuron graphs. Polysemanticity, the phenomenon where a neuron exhibits multiple unrelated behaviours, is one of the current major challenges of neuron interpretability \citep{elhage2022toy}. Interestingly, when present, polysemanticity often shows up clearly in the neuron graphs as distinct, disconnected subgraphs. For example, in Figure \ref{figure:polysemantic}, there are three separate subgraphs corresponding to three clearly distinct behaviours. The top subgraph responds to a phrase in Dutch - variations on \emph{de betrokken}, where not all tokens in \emph{betrokken} were important enough to include in the graph. The middle subgraph responds to a phrase in English - vartions on \emph{a fun, over the top}. The bottom subgraph responds to a phrase in Swedish - \emph{kollegers berättigade}, with unimportant tokens not included. This natural separation of behaviours into separate subgraphs could potentially make it easier to interpret polysemantic neurons, but more experimentation would be needed to develop and test this further.

\begin{figure}[t]
\centering
\includegraphics[width=7cm]{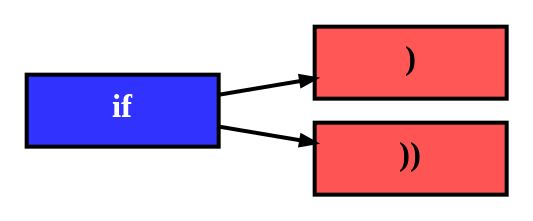}
\caption{A neuron graph representing the syntax for \textit{if} statements with closing brackets.}
\label{figure:code_neuron}
\end{figure}

N2G may work particularly well for neurons that respond to structured text such as code. For example, Figure \ref{figure:code_neuron} shows a graph for a neuron that responds to the syntax for \textit{if} statements containing closing brackets. The current implementation of N2G appears to work better for this \emph {syntactic} neurons than for more complex \emph{semantic} neurons, suggesting it might be a useful tool for interpreting code models, and also highlighting the room for future development to enhance the ability to capture complex semantic behaviour as well.

\begin{figure}[t]
\centering
\includegraphics[width=13.8cm]{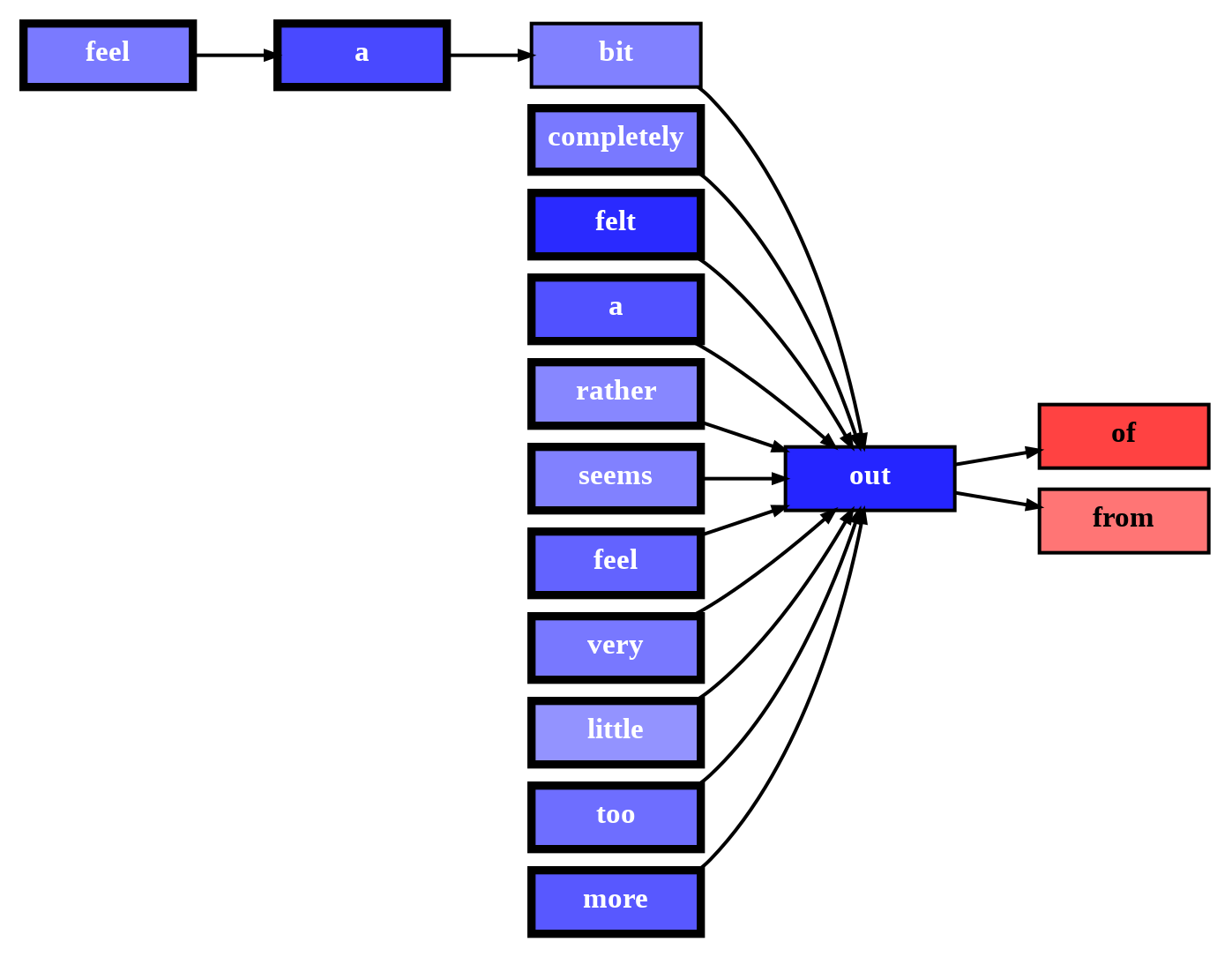}
\caption{A neuron graph with the core token \emph{out} that creates a bottleneck in the graph, as it must always be present for neuron activation despite not being an activating token itself.}
\label{figure:bottleneck}
\vspace*{4in}
\end{figure}

Another interesting behaviour that shows up in some graphs is the presence of a \emph{core} token on which the neuron does not activate. This core token is essential for neuron activation and appears as a bottleneck in the graph, where a variety of nodes in the prior context may cause activation, and there may be multiple activating nodes, that all connect via the core token. For example, in Figure \ref{figure:bottleneck}, \emph{out} is the core token which creates the bottleneck in the graph, with many possible prior tokens and two possible activating tokens. The appearance of this behaviour in neuron graphs suggests that they could be useful for discovering common categories or overall structures of neuron behaviour.

\end{document}

%% file: math_commands.tex
\usepackage{amsmath,amsfonts,bm}

\def\eqref#1{equation~\ref{#1}}

\def\1{\bm{1}}

\DeclareMathAlphabet{\mathsfit}{\encodingdefault}{\sfdefault}{m}{sl}
\SetMathAlphabet{\mathsfit}{bold}{\encodingdefault}{\sfdefault}{bx}{n}

